\documentclass[conference]{IEEEtran}
\usepackage{times}

\usepackage[numbers]{natbib}
\usepackage{multicol}
\usepackage[bookmarks=true]{hyperref}
\usepackage{graphicx}
\usepackage{picinpar}
\usepackage{amsmath}
\usepackage{url}
\usepackage[latin1]{inputenc}
\usepackage{colortbl}
\usepackage{soul}
\usepackage{multirow}
\usepackage{pifont}
\usepackage{color}
\usepackage[dvipsnames]{xcolor}
\usepackage{alltt}
\usepackage{hyperref}
\usepackage{enumerate}
\usepackage{siunitx}
\usepackage{breakurl}
\usepackage{epstopdf}
\usepackage{pbox}

\usepackage[ruled,linesnumbered, lined, noend]{algorithm2e}
\usepackage{booktabs}
 
\usepackage{bm}
\usepackage{hyperref}
\usepackage{amssymb}
\usepackage{romannum}
\usepackage{bbm}
\usepackage[caption=false,font=footnotesize]{subfig}

\usepackage[normalem]{ulem}

\hypersetup{hidelinks} 
\usepackage[switch]{lineno}

\usepackage{siunitx}

\begin{document}

\title{RSS Pioneers:\\
Enhancing Deformable Object Manipulation By Using Interactive Perception and Assistive Tools
}

\author{
Peng Zhou\\
The University of Hong Kong
}

\maketitle



\IEEEpeerreviewmaketitle

In robotic manipulation research \cite{billard2019trends, mason2018toward, yin2021modeling, sanchez2018robotic}, the performance of deformable object manipulation remains a lower level than what humans can achieve due to the inherent properties of deformable objects.
First, deformable objects have infinite degrees of freedom, which makes them non-trivial to perceive and estimate their state configurations. This is in contrast to states of rigid objects (that are typically denoted by fixed and low-dimensional representations).
Second, deformable objects possess complex dynamics, which makes it difficult to predict the future configuration or representation of the deformable object after some actions executed.

Despite the recent focus on deformable objects in robotic manipulation \cite{lin2020softgym, seita2021learning, antonova2021dynamic, zhou2021lasesom, zhu2022challenges, chen2022diffsrl, zhou4432733human}, most of the approaches rely on a static camera and simple empty-handed manipulation (e.g., picking-and-placing, pushing, and so on), which restricts the task performance of deformable object manipulation (DOM) at a much lower level than humans can. 
One possible solution to enhance the performance of manipulating deformable objects is to break these two assumptions: static vision and empty-handed manipulation, introducing interactive perception and assistive tools.
Our first insight is that during deformable object manipulation there exits optimal or sub optimal perspectives to observe and estimate the states of the deformable objects. By adjusting the perception view, it would be much easier for the state estimator or representation module to determine the configuration with a high belief. 
Furthermore, by exploring and exploiting the \textit{action perception regularity} \cite{bohg2017interactive} between perception module and the robot manipulator, the approach of interactive perception is able to create promising interaction of \textit{perception for better manipulation} and \textit{manipulation for better perception}.
Our second insight is to utilize assistive tools to enhance the manipulation performance.
The ability to use tools is one of the hallmarks of human intelligence \cite{jamone2022modelling}. 	
If robots could use tools as flexibly as humans do, they could operate successfully in unstructured environments and solve a large number of tasks that are currently outside of their capabilities \cite{tee2022framework}.
For example, by introducing a folding board for robotic garment folding task, the robot can accomplish the folding with high efficiency and effectiveness.
That is because the assistive tool is able to minimize object deformation into a reduced configuration space, thus better handling the complex dynamics of garments compared to previous non-tool-based approaches.
As a result, our aim in this research statement is to solve the deformable object manipulation problem by introducing \textbf{interactive perception} (IP) and \textbf{assistive tools} to further enhance the manipulation performance on deformable objects. 

\begin{figure}[htbp]
\centering
\includegraphics[width=1\columnwidth]{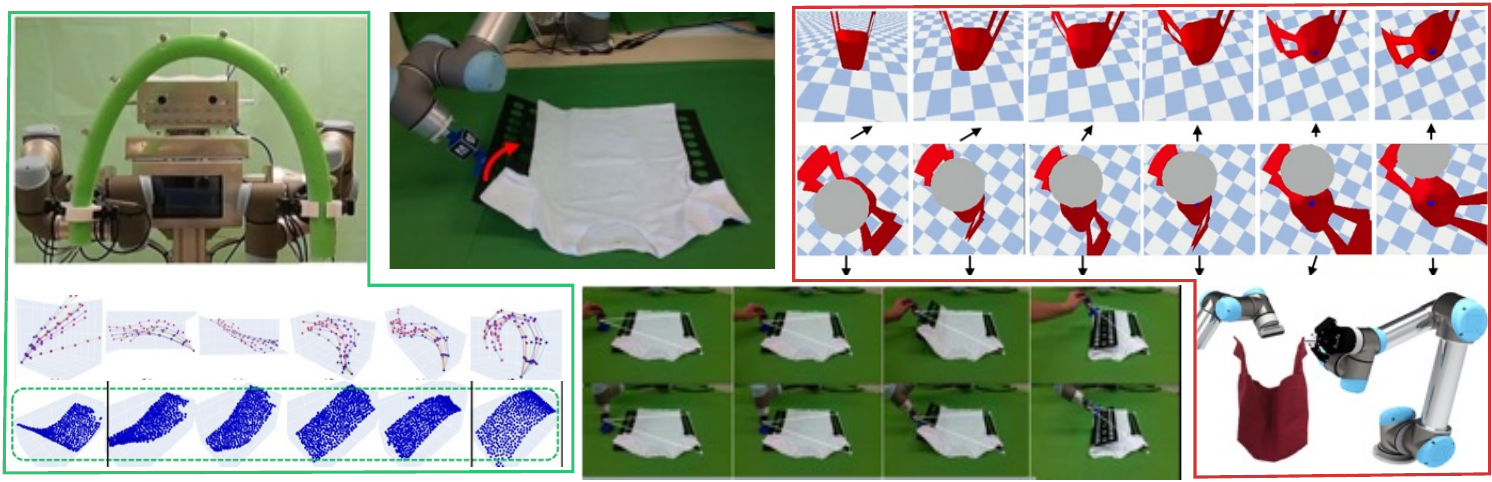}
\caption{
An overview of selected previous work on deformable object manipulation.
From left-to-right: shape planning for deformable objects using bimanual manipulation \cite{zhou2021lasesom}, tool-based garment folding \cite{zhou2022folding}, and performing bag manipulation with active perception in Pybullet \cite{coumans2021} simulation.
}
\label{fig_overview}
\vspace{-0.4cm}
\end{figure}

\subsection{Interactive Perception for Bag Manipulation}
Interactive Perception exploits various types of forceful interactions with the environment to simplify and enhance perception \citep{bohg2017interactive}. IP has been shown effective in exploring and exploiting object and environment properties, with the main focus on rigid and articulated objects \citep{cheng2018reinforcement,novkovic2020object,martin2014online, qi2021contour}.
Arbitrary deformable objects add to the complexity of perception and manipulation problems due to their complex dynamics and infinite-dimensional configuration spaces~\citep{yin2021modeling}. Despite the recent focus on deformable objects in robotic manipulation, most of the approaches rely on a static camera~\citep{li2018learning, zhou2021lasesom}. 

In this project, we develop the first IP framework considering deformable object manipulation (DOM) with an active camera.
We base our approach on two agents, a perceiver and an actor, that act in coordination to allow a more effective execution of the underlying task involving a deformable object. 
An example is shown in Fig.~\ref{fig_overview}(c): the pose of the camera is controlled to provide a view that allows for a better observation of the object in the bag, while the bag is being opened. The proposed methodology follows the idea of Action Perception Regularity (APR) \cite{bohg2017interactive}, stating that the interaction reveals regularities in the combined space of sensor information $(\mathcal{S})$ and action parameters $(\mathcal{A})$ over time $(t)$. Thus, despite the high dimensionality of the integrated space ($\mathcal{S}\times \mathcal{A} \times t$), the signal represented in this space exhibits more structure if properly modeled. 
Our approach models the regularity/structure of the integrated perception-action space in a partially observable Markov Decision Process (POMDP) context. 
The key to implement this APR is constructing a subspace of $\mathcal{A}$ for efficient action search and maximizing the accumulated reward. 
 Specifically, we introduce a manifold-with-boundary formulation to represent an efficient camera action subspace and explicitly encode the coupling between the perceiver, deformable object and the actor. We provide detailed algorithms to calculate the manifold and integrate it into a general action search framework such as reinforcement learning (RL), commonly used in the current state of the art. 
 In summary, this work contributes a formulation involving both active camera and a deformable object, which to our knowledge is the first in the context of interactive perception. We show that the formulation can be instantiated with common model assumptions and tractable computation on manifolds. We also compare the performance of algorithms to the state of the art method, demonstrating a significant performance boost in challenging DOM scenarios. 

%

\subsection{Assistive Tools for Garment Manipulation}
Garment folding is a clear example of a monotonous activities of daily living (ADL) task that can theoretically be performed by robots but which, in practice, is difficult to solve by using these state-of-the-art strategies \cite{borras2020grasping, jangir2020dynamic}.
One possible solution to alleviate the complexity of manipulating fabrics is to enable the robot to learn how to use an assistive tool by observing an expert demonstration and then imitating the behavior (referred to as imitation learning (IL) \cite{schaal1999imitation, argall2009survey}).
Our aim in this work is to solve the garment folding problem by using an assistive tool under the imitation learning paradigm.

Previous works \cite{huo2022keypoint} rely either on computationally expensive representation algorithms (i.e., based on polygonal models \cite{stria2014garment}, particle-based models \cite{hou2017particle}) or elaborately designed manipulation pipelines for ad-hoc cases of study \cite{doumanoglou2016folding}.
Furthermore, traditional IL approaches \cite{hussein2017imitation} assume that the state is fully observed (which is very difficult to satisfy in practice) and utilize model-free learning approaches (which typically demand large amounts of sample data).
As classical imitation policies directly generate a mapping from the high-dimensional image space to the action space, they lack physical interpretability.

Our aim in this work is to use imitation from observations \cite{liu2018imitation} for tool-based garment folding, with a policy that must be learned from observations of a demonstrator's state transitions only.
Therefore, model-based learning techniques \cite{moerland2020model} are needed to learn the underlying dynamics model by collecting action information.
Though inverse reinforcement learning (IRL) \cite{arora2021survey} can solve imitation learning from observations, it typically learns the policy from a high-dimensional observation space (e.g., images), which has proven to be difficult to apply in practice and lacks interpretability of the learned policy.
To deal with this, we seek to encode the dynamics model equation into a low-dimensional representation to help learn imitation policies with high interpretability.
Compared with model-free approaches \cite{ramirez2022model}, model-based learning requires fewer interactions with the environment, thus, making policy learning more sample-efficient \cite{chebotar2017combining}; These learned models could be easily applied to new tasks \cite{englert2013model}.

In this project, we formulate our proposed solution as a POMDP and employ a view-invariant dense descriptor model for detecting visual correspondences of the manipulated object (that is an assistive tool for garment folding); By introducing an assistive folding board, both manipulation efficiency and folding task quality are guaranteed.
More importantly, given that the predefined points on the folding board can largely reduce state dimensions and object deformability, this strategy enables the encoding of the state observation into a high-level (but low-dimensional) graph-structured representation, here called a \textit{hand-object graph} (HoG) model. 
The vertices of the HoG denote the corresponding key points of the hand and the manipulated object (viz., a folding board), whereas the edges represent their relative 3D spatial configuration. 
We use a graph neural network to build a forward dynamics model over the HoG to predict transitions between HoGs and robot actions. 
During the one-shot imitation, for each time step, the learned dynamics model is used to optimize robot actions as the constraint for a model predictive controller that solves the task with a folding board in a closed-loop manner.
The original contributions of this work are summarized as follows:
\begin{itemize}
	\item A new approach of introducing an assistive folding board to improve the performance of robotic garment folding tasks.
 	\item A novel \textit{hand-object graph} for high-level representation of  robotic tool manipulation tasks. 
	\item A new framework of imitation learning from a single observation based on graph-based forward dynamics model.
\end{itemize}


\subsection{Future Directions}
In my future research, I intend to improve deformable manipulation by incorporating reactive manipulation \cite{zhou2023neural}, human robot collaboration, and 
interactive perception for tool-based deformable object manipulation tasks.

\textbf{Reactive Manipulation of Deformable Objects.}
Many studies solve deformable object manipulation with shape servoing techniques, ending up with a target shape configuration that remains unchanged.
In many cases, our target shape must be altered to meet various requirements. For example, a robotic system must manipulate the ring of the bag into different configurations when packaging objects of different shapes.

\textbf{Human Robot Collaboration for Deformable Object Manipulation.}
Today, robotic systems have demonstrated promising results in human-robot collaboration on rigid objects. 
The current research on manipulating deformable objects does not involve human actions and merely utilizes human collaboration to perform tasks on deformable objects, such as robotic bed-making with a human assistant.
As a result, it is imperative to take into account both human intention and the feedback signal provided by the manipulable object. 
Due to this, it is crucial to reconsider the deformable object problem in the context of human-robot collaboration.

\textbf{
Interactive Perception for Deformable Object Manipulation.
}
To manipulate a deformable object in unstructured environments, robots need to find a solution to estimate its current shape. However, tracking the shape of a deformable object can be challenging because of the object's high flexibility, (self-)occlusion, and interaction with obstacles. Building a high-fidelity physics simulation to aid in tracking is difficult for novel environments. Instead we focus on introducing interactive perception for better coupling between perceptions and robotic actions.

\clearpage

\bibliographystyle{IEEEtran}
\bibliography{refs}

\end{document}